\documentclass[runningheads]{llncs}
\usepackage[T1]{fontenc}
\UseRawInputEncoding

\usepackage{booktabs}
\usepackage{multirow}

\usepackage{xcolor}
\usepackage{colortbl}

\usepackage{graphicx,verbatim}
\usepackage{url}
\usepackage{color}

\usepackage{amssymb}
\usepackage{makecell}
\usepackage[table]{xcolor}
\usepackage{xcolor}
\usepackage{array}
\usepackage{amsmath}
\usepackage{float}
\usepackage{hyperref}
\usepackage[dvipsnames]{xcolor}

\newcommand \kkred[1]{\textcolor{Red}{#1}}  
\newcommand \kkgre[1]{\textcolor{ForestGreen}{#1}}

\begin{document}

\title{Modality-Agnostic Input Channels Enable Segmentation of Brain lesions in Multimodal MRI with Sequences Unavailable During Training}

\titlerunning{Modality-Agnostic Input Channels}

\author{Anthony P. Addison\inst{1}\orcidID{0009-0008-4938-6123} \and
Felix Wagner\inst{1}\orcidID{0009-0004-6683-171X} \and
Wentian Xu\inst{1}\orcidID{0009-0009-2440-007X} \and
Natalie Voets\inst{2}\orcidID{0000-0001-8078-4471} \and
Konstantinos Kamnitsas\inst{1}\orcidID{0000-0003-3281-6509}}
\authorrunning{A.P.Addison et al.}
%
\institute{Department of Engineering Science, University of Oxford, Oxford, UK \and Nuffield Department of Clinical Neurosciences, University of Oxford, Oxford, UK \\
\email{anthony.addison@eng.ox.ac.uk}}

\maketitle              
\begin{abstract}
Segmentation models are important tools for the detection and analysis of lesions in brain MRI. Depending on the type of brain pathology that is imaged, MRI scanners can acquire multiple, different image modalities (contrasts). 
\
\
Most segmentation models for multimodal brain MRI are restricted to fixed modalities and cannot effectively process new ones at inference. Some models generalize to unseen modalities but may lose discriminative modality-specific information.
\
This work aims to develop a model that can perform inference on data that contain image modalities unseen during training, previously seen modalities, and heterogeneous combinations of both, thus allowing a user to utilize any available imaging modalities.
\
We demonstrate this is possible with a simple, thus practical alteration to the U-net architecture, by integrating a modality-agnostic input channel or pathway, alongside modality-specific input channels.
\
To train this modality-agnostic component, we develop an image augmentation scheme that synthesizes artificial MRI modalities. Augmentations differentially alter the appearance of pathological and healthy brain tissue to create artificial contrasts between them while maintaining realistic anatomical integrity. We evaluate the method using 8 MRI databases that include 5 types of pathologies (stroke, tumours, traumatic brain injury, multiple sclerosis and white matter hyperintensities) and 8 modalities (T1, T1+contrast, T2, PD, SWI, DWI, ADC and FLAIR). The results demonstrate that the approach preserves the ability to effectively process MRI modalities encountered during training, while being able to process new, unseen modalities to improve its segmentation. Project code: \textit{https://github.com/Anthony-P-Addison/AGN-MOD-SEG}


\end{abstract}
\section{Introduction}

Magnetic resonance imaging (MRI)  is an invaluable tool for the diagnosis and analysis of lesions. MRI scanners can be configured with different pulse sequences to acquire varying contrasts between tissues. A contrast setting defines an MRI "modality". 
\
Different MRI modalities provide enhanced visualization of various brain tissues and lesions. Thus \emph{multimodal MRI} uses a set of modalities for comprehensive brain analysis. Depending on the pathology studied and the clinical practice in the specific imaging center, a different set of MRI modalities are acquired for the patient.

Neural networks are effective models for the segmentation of brain lesions \cite{Isensee2024NnUNetSegmentation,Kamnitsas2017EfficientSegmentation,Chen2019S3D-UNET:Segmentation}. These models are typically trained for one type of lesion at a time,  using a predefined set of MRI modalities. Therefore, they cannot process a different set of modalities, or a new modality that has not been seen in the training data. Given that different sets of modalities are acquired at different imaging centers, even for the study of the same type of lesion, we need more flexible tools for multimodal brain MRI segmentation. 
Therefore, the objective of this work is to develop a multimodal model that can segment brain lesions using any MRI modalities available to the user, even types that have not been encountered in the training data. A significant amount of work has been done to develop methods that can handle \textbf{missing modalities} at inference time \cite{Azad2022SMU-Net:Modalities,Havaei2016HeMIS:Segmentation}. 
However, these methods cannot deal with new modalities at inference time. 
\
\textbf{Domain adaptation} methods adapt a model trained on a source data domain to effectively generalize across a new target data domain. This approach was initially demonstrated using adversarial networks to adapt to a target set of MRI modalities \cite{Kamnitsas2017UDA}. Fine-tuning has also been applied to adapt to new brain MRI datasets \cite{Ghafoorian2017TransferSegmentation,Valverde2019One-shotNetworks}. 
The drawback of these methods is the requirement to pre-collect data from the target domain to retrain (adapt) model weights, which can be time-consuming, resource-intensive, and often impractical. Instead, here, our aim is to develop models that better generalize to new modalities directly.
\
\textbf{Domain Generalization} (DG) aims to train a model that generalizes to unseen domains without further adaptation, such as through contrastive learning \cite{dou2019domain} or non-linear intensity augmentations \cite {OuyangCausality-inspiredSegmentation,Billot2023SynthSeg:Retraining}. In principle, such DG methods learn generic, modality-agnostic representations, but may compromise model ability to extract modality-specific, discriminative representations, useful in multimodal MRI processing. Our work alleviates this by integrating modality-agnostic filters, trained via augmentations appropriate for modality generalization, into a model that preserves modality-specific filters.
Other recent related work develop \textbf{foundation models} with MedSAM \cite{Ma2024SegmentImages} as a representative example. These are trained on diverse data with different modalities (CT, MRI, XRay, etc) and pathologies, often with the aim of segmenting any type of medical image. However, these exhibit inconsistent performance on unseen domains \cite{Akram2023Citation:Segmentation}. In addition, they are trained by processing only one input modality at a time, regardless of its type, to learn modality-agnostic input filters. They are thus unable to process multimodal MRIs - the objective of this work. The first step towards foundational models for multimodal MRI segmentation of brain lesions was demonstrated recently \cite{Xu2024FeasibilitySegmentation}, by training a U-net on multiple databases with diverse sets of modalities. Other works train a mixture-of-experts for this task, rather than a single model with generic foundational knowledge \cite{zhang2024foundation}. These works cannot handle new modalities at inference time. We build upon the work of \cite{Xu2024FeasibilitySegmentation} to alleviate this limitation.

In this context, the primary contributions of this work are as follows:

 \begin{itemize}
    \item  We develop a model for segmentation of multiple types of lesions in multi-modal MRI, which is capable of making predictions using a modality unavailable during training, along with modalities encountered during training. For this, we developed an architecture that integrates a modality-agnostic input channel or pathway into a multi-modal model with modality-specific filters.
    \item We present a training framework for such a model with modality-agnostic components, based on Modality Dropout and augmentation strategies that synthesize artificial contrasts between brain and lesions.
    \item Experiments using 8 MRI databases of brain lesions demonstrate that the model preserves the ability to process MRI modalities seen during training while being able to process new modalities previously unseen at inference time to improve its segmentation.
 \end{itemize}

\section{Method}

We first introduce a method to train a model on MRI scans from multiple datasets each with different sets of input MRI modalities and pathologies with modality specific input channels. Then, we introduce our method to enable the model to process modalities not seen during training when evaluated on new datasets with unseen modalities. 

\subsection{Modality Specific Input Channels}
\label{sec:modality_specific_input_channels}

\noindent A version of a U-Net \cite{Zhou2018Unet++:Segmentation} model was implemented as the backbone for all experiments. The total number \(\mathcal{N}\) of training databases utilized \(\mathcal{D} = \{\mathcal{D}_{1}, \ldots, \mathcal{D_{N}}\}\), where each database \(\mathcal{D}_i \in \mathcal{D}\) has a unique set of modality combinations \(\mathcal{M}_{i}\), and \(\mathcal{C}_{i}=|\mathcal{M}_i|\) is the number of unique channels utilized per database. 
The number of modality \textbf{specific} input channels of the U-Net is set to \(\ \mathcal{C}_{sp} = |\mathcal{M}_{1}\cup \ldots \cup \mathcal{M_{N}}|\), which is the number of \textit{unique} modalities across all \(\mathcal{N}\) databases. 
Each modality has a dedicated input channel, enabling the U-Net to learn modality-specific features.
When a modality is absent in a given case, its corresponding input channel is filled with zeros. This model is referred to as the {\bfseries Standard} model configuration (Eq.~\ref{eq:standard}). The model is based on \cite{Xu2024FeasibilitySegmentation}.

\begin{figure}
\centering
\includegraphics[scale = 0.30]{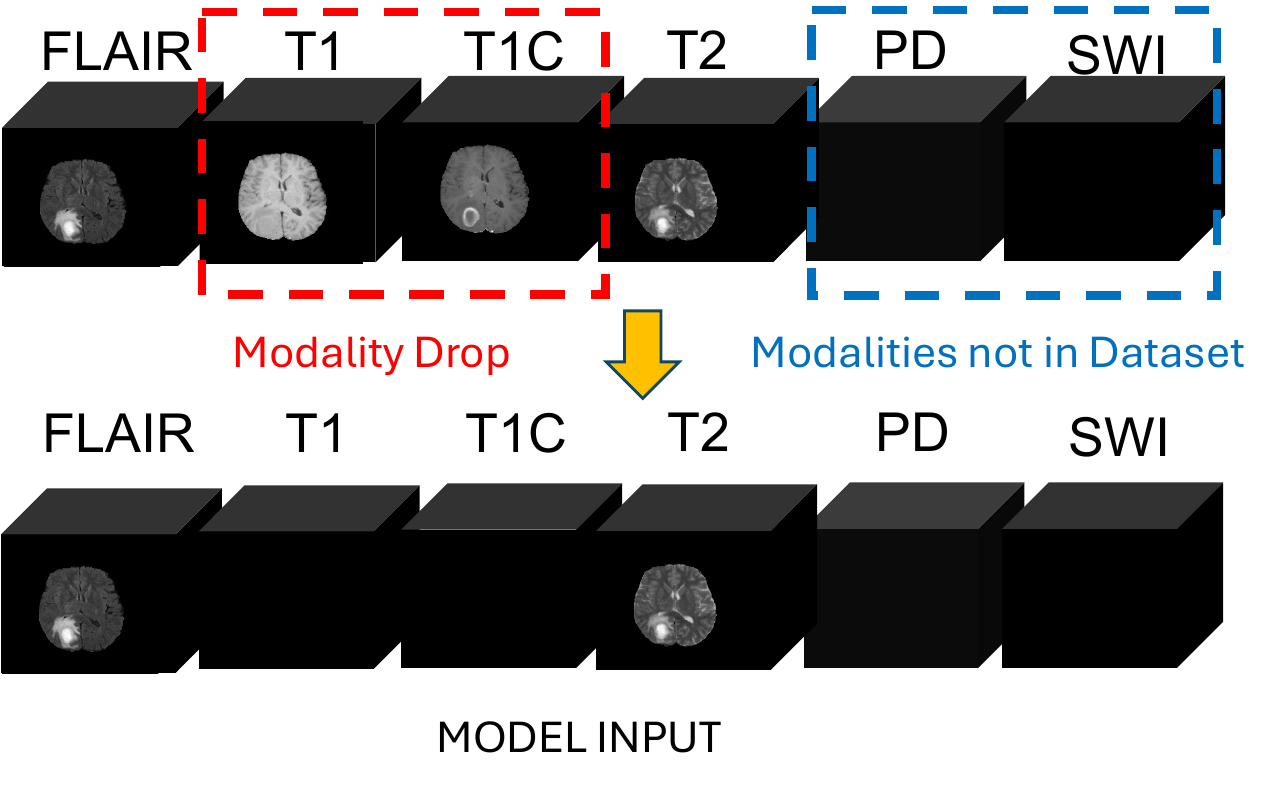}
\caption{ Modalities not in dataset and modalities dropped are set to zero. The resulting model input consist of the retained modalities (Example from BRATS).}
\label{fig_dropout}
\end{figure}

\noindent \textbf{Modality Dropout During Training:} 
A model learns to associate predefined sets of modalities with specific lesions, limiting its ability to generalize to cases where a subset of modalities is missing at inference time. To encourage the model to learn representations that are robust to input modality combinations not seen during training, we employ Modality Dropout following \cite{Xu2024FeasibilitySegmentation}. Using Modality Dropout during training, random input modalities are filled with zeros, optimizing the model to segment regardless of the modalities available (Fig. \ref{fig_dropout}).

\begin{figure}[h]
\includegraphics[width=\textwidth]{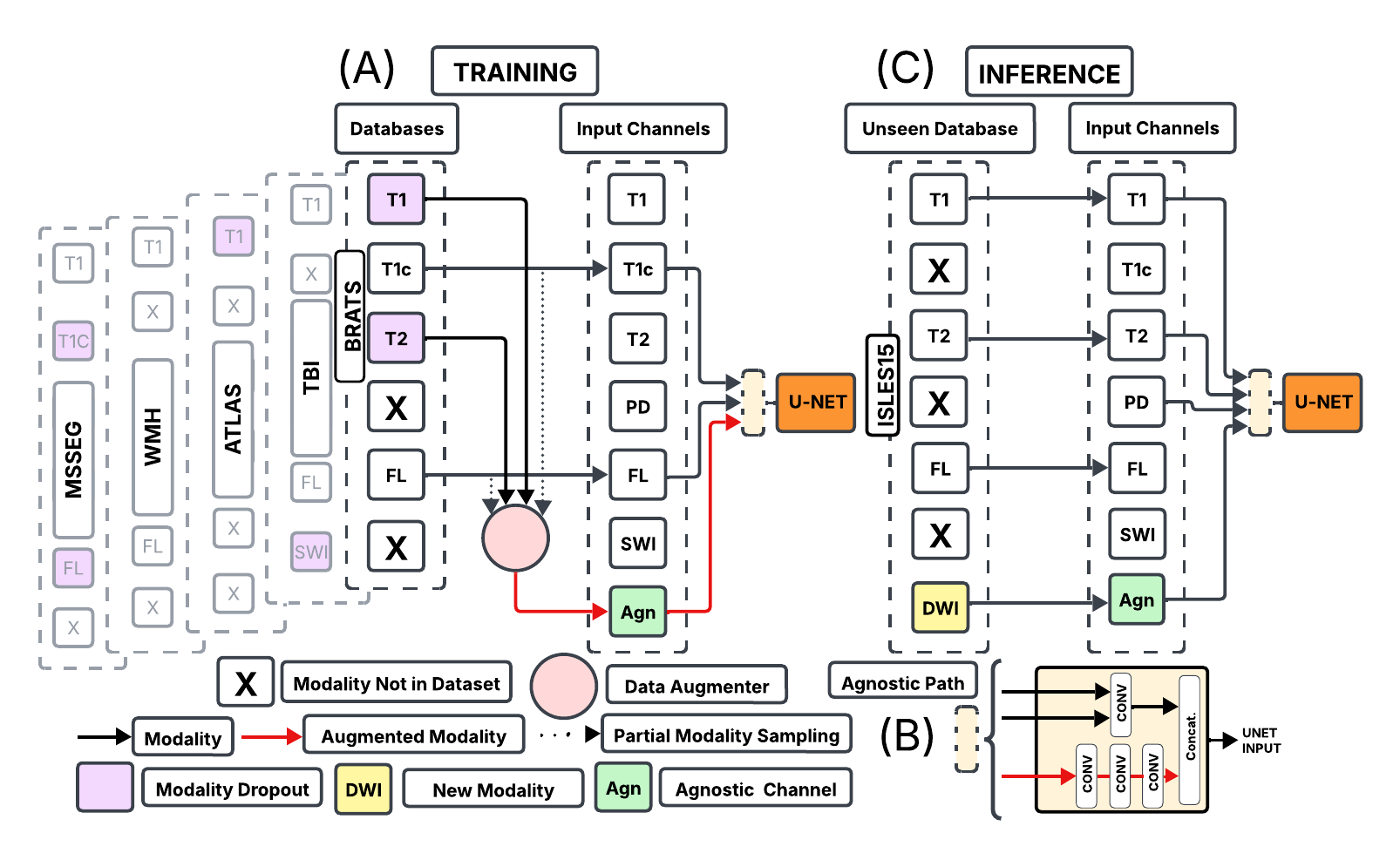}
\caption{ Overview of the modality input configuration decision architecture. {\bfseries (A) Training:} For each case in a dataset, input channels corresponding to modalities that are either absent from the dataset or randomly dropped are set to zero. Dropped modalities, as well as selected components of the retained modalities, are augmented using the data augmenter. The augmented modality is inserted into the agnostic channel, while the original modalities remain in their respective modality-specific channels. {\bfseries(B) Agnostic Pathway:} optional block where the agnostic channel input is processed initially and then concatenated with the other modalities. {\bfseries (C) Inference:} a held out dataset with a combination of seen and unseen modalities is tested. New modalities are placed in the agnostic channel.}
\label{fig6}
\end{figure}

\subsection{\bfseries Modality Specific and Agnostic Input Channels}
{\bfseries  Agnostic Channel:} To handle an unseen modality at inference time, an additional modality-agnostic channel \(\mathcal{C}_{agn}\) is added to the channel set \(\mathcal{C}_{sp}\), resulting in \(\mathcal{C}_{sp+agn} = \mathcal{C}_{sp} \cup  \{\mathcal{C}_{agn}\}\). The extra channel provides a dedicated space to process a new modality during inference. During training, dropped modalities \(\mathcal{M}_{drop}\) and augmented modalities \(\mathcal{M}_{aug}\) are used as inputs to this channel (the following paragraph outlines the augmentation method). The model architecture remains unchanged relative to the standard model, except of one additional input channel. The model with modality-agnostic channel is formulated in Eq.~\ref{eq:invar_channel}.

\noindent \textbf {Agnostic Path:} To increase the model's representational capacity to process an additional input modality, we add an additional pathway prior to the U-Net architecture to process this input modality separately. The pathway consists of 3 convolutional layers, where the first layer processes a 1 channel input and the final convolutional layer outputs 8 feature maps.
The output of these layers is concatenated with features of the inputs of the modality-specific channels, generated by one convolutional layer that preserves the number of input channels (Fig.~\ref{fig6}(b)). The agnostic path model is provided in Eq.~\eqref{eq:invar_path} where, \(\varphi_{\text{path}}\) represents the agnostic pathway. 

\begin{align}
f_{\text{standard}}(\mathcal{M}_{sp}) &= \text{U-Net}(\mathcal{M}_{sp}) 
\label{eq:standard}
\\
f_{\text{agnostic-channel}}(\mathcal{M}_{sp}, \mathcal{M}_{aug} ) &= \text{U-Net}(\mathcal{M}_{sp} \oplus \mathcal{M}_{aug} )
\label{eq:invar_channel}
\\
f_{\text{agnostic-path}}(\mathcal{M}_{sp},\mathcal{M}_{aug}) &= 
\text{U-Net}(\text{conv} (\mathcal{M}_{sp}) \oplus \varphi_{\text{path}}(\mathcal{M}_{aug})))
\label{eq:invar_path}
\end{align}

\begin{figure}[h]
\includegraphics[width=\textwidth]{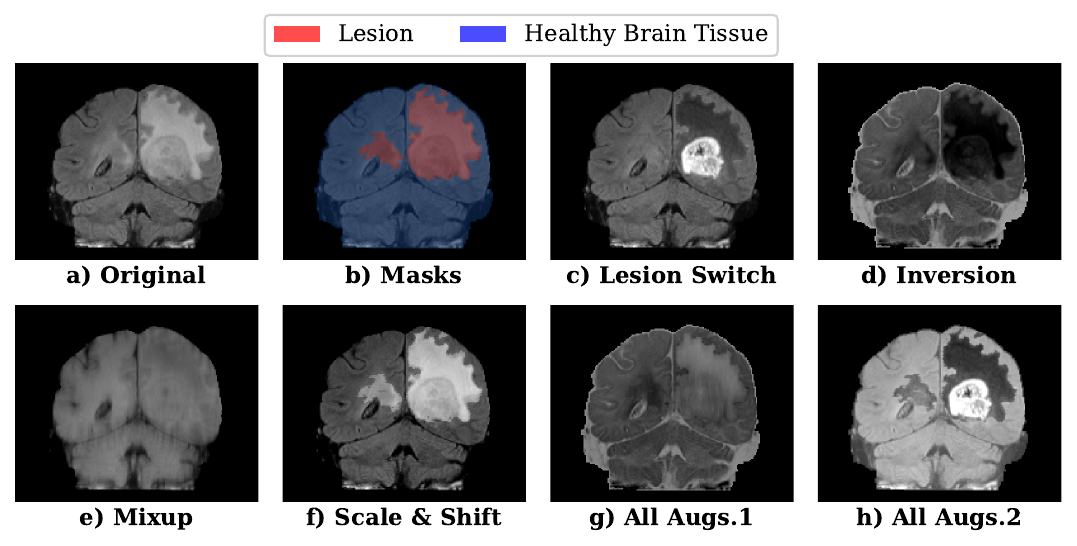}
\caption{Spatial contrast augmentation techniques for brain MRI. Representative examples of methods applied to a case from BRATS \cite{Bakas2017AdvancingFeatures}. All Augs.1 and 2 are examples where all augmentations are included. These techniques enhance model robustness by simulating realistic intensity variations while preserving anatomical structure.} 
\label{fig:augmentations}
\end{figure}

\noindent\textbf{Training with Augmented Modalities:} To enable the added modality-agno\- stic input filters to process unseen modalities, we generate augmented modalities from the available ones. Augmentations are designed with the aim of preserving structure while producing a wide range of synthetic modalities. At each training iteration, we create augmented modalities primarily exploiting images dropped by Modality Dropout as they are not present in modality-specific channels for the current iteration. 
Assigning the same modality both to a modality-specific and the modality-agnostic input channels creates redundancy.
\
In addition, certain augmentations we , use parts of the retained modalities (Lesion Switch, MixUp). Each enhancement is applied to the lesion and the healthy brain regions with independently calculated probabilities \(p \in [0,1]\) for each sample. The lesion and brain masks are used to guide region-specific transformations. The following augmentation techniques are used: {\bfseries Shift and Scale:} of pixel value: \( \mathcal{M}_{aug} ^\prime = \alpha\ \mathcal{M}_{aug}+ \beta\ \) where the scale and shift parameters are sampled as follows: \(
\{\alpha, \beta\} \sim \{\mathcal{U}(a, b), \mathcal{U}(c, d)\}
\), where \( (a, b) \) denotes the upper and lower bounds of the shift range and \( (c, d) \) denotes the upper and lower bounds of the scale range. 
{\bfseries Lesion Switch:} Produces an augmented modality by using the lesion mask to insert the lesion region from one modality into the healthy brain tissue from another modality. {\bfseries Inversion:} inverting the pixel values \( \mathcal{M}_{aug} ^\prime = - \mathcal{M}_{aug}\). {\bfseries MixUp \cite{ZhangMixup:MINIMIZATION}:} Method based on linear interpolation between two modalities \({\mathcal{M}_{aug}}^\prime = \lambda\ {m}_{1} + (1-\lambda){m}_{2}\), where \({m}_1\) and \({m}_2\) are modalities of the processed sample (e.g. T1 and T2) and \(\lambda \in [0,1]\) is the parameter that controls the weighting. Fig.~\ref{fig:augmentations} shows augmentation examples.

All augmentations are applied with independent probabilities sequentially in order to create a new synthetic modality \(\mathcal{M}_{aug}\). We applied a specific order of implementing the augmentations as improper ordering can result in unrealistic synthetic data or instability during training: {\bfseries Lesion Switch (only for lesion augmentations) $\boldsymbol{\rightarrow}$ Inversion  $\boldsymbol{\rightarrow}$ MixUp  $\boldsymbol{\rightarrow}$ Scale $\boldsymbol{\rightarrow}$ Shift}. Formally, this can be expressed as:

\begin{equation}
M_{aug} = (A_{s_1}^{r_1} \circ A_{s_2}^{r_2}\circ \ldots \circ A_{s_k}^{r_k} ) ( M_{drop} )
\end{equation}

where:
\begin{itemize}
\item 
\(
S = (s_1, s_2, \ldots, s_k), \text{where}, (s_1, s_2, \ldots, s_k) \text{ is a subsequence of: } \) \\ (Lesion Switch, Inversion, MixUp, Scale, Shift)
\
\item The number of augmentations applied, \( k \), is given by: \(k = |S|  \quad \text{where} \quad k \in \{0,1, 2, 3, 4, 5\}\).
\item $r \in \{\text{lesion}, \text{brain}, \text{uniform}\}$ specifies the region of application.
\item $A_{s_i}^{(r)}$ applies the $s_i$-th augmentation to region $r$ only.
\item $M_{\text{drop}}$ is the dropped modality and $M_{\text{aug}}$ is the final augmented modality.
\end{itemize}

\section{Experiments}

\subsection{Datasets and Training Details}

{\bfseries Datasets:} Eight 3D brain MRI datasets were utilized in this study. They are as follows: tumour cases from BRATS 2016 \cite{Bakas2017AdvancingFeatures}, multiple sclerosis cases from MSSEG 2016 \cite{Commowick2018ObjectiveInfrastructure}, stroke lesion cases from ATLAS v2.0 \cite{Liew2022AAlgorithms}, white matter hyperintensity cases from the WMH challenge \cite{Kuijf2019StandardizedChallenge}, sub-acute ischemic stroke lesion cases from the SISS subtask of ISLES15 2015 \cite{Maier2017ISLESMRI}, and acute stroke lesion cases from ISLES22 \cite{HernandezPetzsche2022ISLESDataset}. Two datasets are private from our institutions, Traumatic Brain Injury(TBI) and TUMOUR2 consisting of tumour cases. Each dataset is characterized by a unique set of modalities and train/test split (Tab.~\ref{database_info}). Pre-processing consisted of skull stripping, resampling images to 1x1x1mm and Z-Score normalization. 

\begin{table}[h]
\renewcommand{\arraystretch}{1.2} 

\caption{Brain MRI datasets used in this study. Modalities present for each dataset indicated by a check mark. Train/Test split of data provided.} \label{database_info}
\centering
\resizebox{0.8\textwidth}{!}{
\begin{tabular}{|c|c|c|c|c|c|c|c|c|c|c|}

\toprule

\multirow{2}{*}{\textbf{Datasets}} & \multicolumn{8}{c|}{\textbf{Modalities}} &\multirow{2}{*}{\textbf{Train/Test}}& \multirow{2}{*}{\textbf{Total Size}}  \\
 \cline{2-9}
 & \textbf{T1} & \textbf{T1C} & \textbf{T2} & \textbf{PD} & \textbf{SWI} & \textbf{DWI} & \textbf{ADC} & \textbf{FLAIR} & & \\

\midrule
BRATS & \checkmark & \checkmark & \checkmark & & & & & \checkmark & 444/40 & 484  \\
ATLAS & \checkmark &  &  & & & & &  & 459/195 & 654  \\
MSSEG & \checkmark & \checkmark & \checkmark & \checkmark & & & & \checkmark & 37/15 & 53  \\
TBI & \checkmark & & \checkmark & & \checkmark & & & \checkmark & 156/125 & 281  \\
WMH & \checkmark & & & & & & & \checkmark & 42/18 & 60  \\
ISLES22 & & & & & & \checkmark & \checkmark & \checkmark & 175/75 & 250  \\
ISLES15 & \checkmark & & \checkmark & & & \checkmark & & \checkmark & 0/28 & 28  \\
TUMOUR2 & \checkmark & & & & & & & & 0/57 & 57  \\

\bottomrule
\end{tabular}
}
\end{table}

\noindent {\bfseries Training Details:} To address database imbalance, all training databases are over sampled to the size of the largest training database. At input if a modality is not present for the current dataset, the corresponding modality input channel/s to the model are zeroed. All different lesion labels are merged into one binary 'lesion' label across all databases. 
\
This study evaluates model performance using two training configurations, each with different unseen datasets and unseen modalities, to assess generalization across dataset and modality combinations. In each configuration, a subset of datasets are used for training, while the remaining databases are held out for evaluation. {\bfseries Train Setting 1:} WMH, MSSEG, BRATS, TBI, ATLAS and {\bfseries Train Setting 2:} ISLES2022, MSSEG, BRATS, TBI, ATLAS (FLAIR removed). 
\
A combined Dice+Cross Entropy loss is implemented for training, with each run consisting of 600 epochs. The learning rate is initially set to \(\times 10^{-4}\) and reduced to \(\times 10^{-5}\) after 350 epochs. All models are trained with a batch size of 2 where each sample is randomly cropped to 96 x 96 x 96 pixels. The same augmentation parameters were applied to both settings. 


\noindent \textbf{Baselines:} To evaluate the proposed Agnostic Channel model and Agnostic Path model, we compare them to the following models (all methods are trained both on Train Setting 1 and 2). {\bfseries Standard Model: (see Sec.~\ref{sec:modality_specific_input_channels})}. 
{\bfseries Single Input Channel:} A model with a single input channel is trained using a random modality per training sample. No cross-modal interactions are learned. At inference time,  the dice score is calculated by averaging predictions made with each available modality separately.
\
{\bfseries Shuffle Input Channels:}  Modalities are randomly shuffled between input channels during training. The dice is the average of three tests of randomly assigned modalities to input channels. Channels do not learn modality-specific features and cross-modal interactions are not effectively learned due to constantly changing the input combinations.

\subsection {Results and Discussion}

\noindent\textbf {Additional Agnostic Channel and Path:} Table~\ref{main_results} shows evaluation of the Agnostic Channel and Agnostic Pathway models and comparison against baselines. The results show that the Agnostic Path model improves performance relative to the Agnostic Channel model for both settings for all test datasets. In Setting 1 (DWI in agnostic channel) the Dice improves by \( 6.4 \% \) for ISLES2022 and improves by \( 1.7 \% \) for ISLES15. For Setting 2 (FLAIR in agnostic channel) it improves by \( 6.2 \% \) for WMH and improves by \( 5.5 \% \) for ISLES15. This effect is likely attributable to the increase in input channels from 1 to 8, which elevates the relative weighting of this pathway and may consequently enhance the model’s attention to that channel input. 
\
For the test datasets there is an improvement in performance of the Agnostic Channel model relative to the standard model. In Setting 1 (DWI in agnostic channel) the Dice improves by \( 4.5 \% \) for ISLES2022 and improves by \( 1.9 \% \) for ISLES15. For Setting 2 (FLAIR in agnostic channel) it improves by \( 29.7 \% \) for WMH and improves by \( 15.3\% \) for ISLES15. This implies that the model is learning to extract a combination of information from the agnostic channel and the modality specific channels. 
\
In addition, the pathway does not compromise performance when processing a held out dataset with only seen modalities present (modalities encountered during training) (TUMOUR2) nor when processing known modalities in data similar to those used in training (held-out from training data). Held out training data for setting 2 when FLAIR is applied at test performs better for all dataset except ISLES22. For setting 1 When tested on ISLES22, the shuffled model achieved the best performance. While the agnostic. Path model performed well on DWI alone 
\( 51 \% \), it failed to learn the combination of FLAIR and DWI \( 35.6\% \) effectively.

\begin{table}[h]
\caption{The average Dice(\%) is reported for both Training Setting 1 and Training Setting 2. Results are presented for the Agnostic Path model (Agn Path) and the Agnostic Channel model (Agn Chan). Each model's performance is shown for scenarios where an \textbf{unseen modality is not used / is used}. Symbol "-" denotes that a scenario was infeasible. Standard model cannot process unseen modality. \fcolorbox{black}{white}{\phantom{xx}} Training Datasets (results from held out data); \fcolorbox{black}{Apricot}{\phantom{xx}} Test Datasets with seen modalities; \fcolorbox{black}{YellowGreen}{\phantom{xx}} Test Dataset with seen and  unseen modalities. \kkgre{green}/\kkred{red} denotes \kkgre{increased}/\kkred{decreased} performance when a model processed at inference a modality unseen during training. Modality-agnostic Path can leverage unseen modalities across most settings.
}

\label{main_results}
\renewcommand{\arraystretch}{1.2} 
\resizebox{\textwidth}{!}{
\begin{tabular}{|l|l|l|l|l|l|l|l|l|l}
\toprule
\multicolumn{1}{|c|}{\multirow{2}{*}{\centering\bfseries{Models}}}
& \multicolumn{8}{c|}{\bfseries{Datasets} }\\

\cmidrule(){2-9}

& \multicolumn{1}{c}{\bfseries BRATS}& \multicolumn{1}{c}{\bfseries MSSEG}&\multicolumn{1}{c}{\bfseries ATLAS}& \multicolumn{1}{c}{\bfseries TBI}& \multicolumn{1}{c}{\bfseries WMH}  & \multicolumn{1}{c}{ {\cellcolor{YellowGreen}{\bfseries ISLES22}}}& \multicolumn{1}{c}{ {\cellcolor{YellowGreen}{\bfseries ISLES15}}}&{\cellcolor{Apricot}{\bfseries TUMOUR2}}\\

\midrule
\rowcolor{gray!30}  
\multicolumn{9}{|c|}{\textbf{{\bfseries Setting 1 (No modality removed during training, new modality in test: DWI)}}}  \\
\midrule
Standard & 91.3/- & 70.8/- & 54.4/- & 58.8/- & 75.6/- & 30.1/-& 56.3/-& 72.8/- \\

Shuffled & 79.5/-  & 53.8/-& 53.9/- & 50.6/- &  67.6/-  & 14.8/\kkgre{40.8}  & 44.4/\kkred{43.7}  & 65.8/-\\

Single & 88.2/- &  48.7/- & 48.5/- & 48.5/- & 61.4/- & 16.9/\kkgre{38.5} &46.5/\kkgre{53.7} &72.7/- \\

Agn Chan & 91.0/-&  70.2/- & 55.0/-& 58.4/- & 75.7/- & 30.7/\kkred{29.2} &56.7/\kkred{56.5} & 74.2/-\\

Agn Path & 91.0/-  &  69.9/-&54.8/-&  59.4/- & 75.8/- & 30.4/\kkgre{35.6}& 55.2/\kkgre{58.2} &73.3/- \\

\midrule
& \multicolumn{1}{c}{\bfseries BRATS}& \multicolumn{1}{c}{\bfseries MSSEG}&\multicolumn{1}{c}{\bfseries ATLAS}& \multicolumn{1}{c}{\bfseries TBI}& \multicolumn{1}{c}{ {\cellcolor{YellowGreen}{\bfseries WMH}}}  & \multicolumn{1}{c}{\bfseries ISLES22}& \multicolumn{1}{c}{ {\cellcolor{YellowGreen}{\bfseries ISLES15}}}&{\cellcolor{Apricot}{\bfseries TUMOUR2}}\\

\midrule
\rowcolor{gray!30}  
\multicolumn{9}{|c|}{\textbf{{\bfseries Setting 2 (FLAIR removed during training, new modality in test: FLAIR)}}} \\
\midrule
Standard & 88.2/- &  59.9/- & 55.2/- & 57.3/- & 27.9/- &74.6/- &36.3/- &76.8/- \\

Shuffled&  80.0/\kkgre{84.6} & 44.0/\kkgre{51.0} & 54.0/- & 46.5/\kkgre{52.6} & 23.9/\kkred{11.4} & 66.4/\kkred{53.9} & 39.3/\kkgre{47.5} &67.1/-\\

Single & 86.0/\kkgre{87.7}  & 58.5/\kkred{58.0} & 53.7/- & 45.9/\kkgre{46.8} & 20.5/\kkgre{33.1} & 68.5/\kkred{59.4} & 39.0/\kkgre{40.5} & 70.1/- \\

Agn Chan& 88.3/88.3 & 60.0/\kkgre{60.3}     & 54.8/-  & 57.3/\kkgre{58.0} & 23.4/\kkgre{51.4} &  73.6/\kkred{73.0} & 44.2/\kkgre{46.1} & 77.3/- \\

Agn Path &  88.3/\kkgre{88.6}  & 60.3/\kkgre{65.2}   & 55.3/- & 56.8/\kkgre{57.3}  & 31.6/\kkgre{57.6} &73.8/\kkred{71.3} & 51.1/\kkgre{51.6} & 76.4/-\\

\bottomrule
\end{tabular}}

\end{table}

\noindent \textbf{Ablation Study of Augmentations} To investigate the effect of each individual augmentation for both tissue-specific and uniformly applied augmentation we conducted an ablation study on Training Setting 2 (Tab.~\ref{augment_ablations_1}). Augmenting the data processed by the modality-agnostic path tends to improve performance relative to no augmentation (No Augs). The various types of augmentations influence performance in different ways across datasets with different modalities (WMH, ISLES15). This is likely due to certain augmentations making the synthesized data more similar or divergent from the new modality available during inference.
Tissue-specific augmentations are effective on WMH, whereas uniform augmentations benefit ISLES15. This shows the proposed method has significant further potential if the design of augmentations is optimized to be even more effective across different settings. In Setting 1, for the agnostic path model when testing on ISLES15 (with DWI), applying tissue-specific augmentations increases the dice from 55.3\% without augmentation to 58.2\% with augmentations.

\begin{table}[h]
\renewcommand{\arraystretch}{0.9}
\caption{Performance (Dice\%) of model with modality-agnostic path in Training Setting 2 when augmentations are: {\bfseries Tissue specific -} augmenting the lesion and healthy tissue independently. {\bfseries Uniform -} augmentations applied uniformly across the image.} 
\label{augment_ablations_1}
\centering
\begin{tabular}{l|cc|cc}
\toprule
\multirow{2}{*}{\textbf{Augmentations}} & \multicolumn{2}{c|}{\textbf{Tissue Specific}} & \multicolumn{2}{c}{\textbf{Uniform}} \\
\cmidrule(){2-5}
   & WMH &ISLES15 & WMH & ISLES15\\
\midrule
No Augs. & 54.2 & 48.8  & 54.2 & 48.8  \\
Scale and shift  & 57.1  & 50.4 & 52.5 & 54.8    \\
Lesion Switch  & 54.3 & 51.2 & N/A & N/A \\
Mix up &  53.8  & 51.0   & 53.2  & 50.1  \\
Invert & 57.3  & 49.7  & 57& 52.7  \\ 
Comb. of augs.  & 57.6 & 51.6  & 57.3  & 57.5 \\
\bottomrule

\end{tabular}
\end{table}

\noindent \textbf{Fine-Tuning:}
The modality-agnostic channel allows the assignment of an unseen modality, along with seen modalities, facilitating effective fine-tuning for datasets with both. To evaluate fine-tuning performance on a new dataset, we utilize pre-trained agnostic path models and fine-tune them with test datasets containing both seen and unseen modalities. For Setting 1, we fine-tune ISLES15 with DWI in the agnostic channel. In Setting 2, we conduct two experiments: first, fine-tuning ISLES15 with FLAIR in the agnostic channel, and second, fine-tuning WMH with FLAIR in the agnostic channel, all experiments are fine-tuned over 600 epochs. A 5-fold cross-validation is implemented for ISLES15 and for WMH the standard train/test split is used. 
\
For both Setting 1 and Setting 2, the dice is better when the pre-trained agnostic pathway model is fine-tuned with a new modality assigned to the channel (see Tab.~\ref{finetune}). This is in contrast to randomly initializing weights of the agnostic channel on the pre-trained standard model (creating the agnostic model) and randomly initializing weights of the agnostic pathway on the pre-trained standard model (creating the agnostic path model). The randomly initialized agnostic pathway performs better than the randomly initialized agnostic channel on its own, suggesting that the pathway increases fine-tuning performance. The results show that the pre trained agnostic pathway model is more effective for fine-tuning to domain-specific data.

\begin{table}[H]
\renewcommand{\arraystretch}{1.2} 
\caption{
Fine-tuning models. {\bfseries Setting 1:} ISLES2015 with DWI assigned to the agnostic channel. {\bfseries Setting 2:} two experiments are conducted: one with WMH using FLAIR assigned to the agnostic channel, and another with ISLES using DWI assigned to the agnostic channel. Other modalities occupy their specific channels. For each entry, the results are: {\bfseries Fine-tune agnostic pathway | Fine-tune standard model with randomly initialized agnostic pathway | Fine-tune standard model with randomly initialized channel}. Best performance for each setting is highlighted in bold.
\label{tab1}}
\centering
\begin{tabular}{l|c|c|c|c|c|c|c|c|c}
\toprule
\label{finetune}
\multirow{2}{*}{\begin{tabular}{@{}c@{}} \bfseries Training \\ \bfseries Datasets \end{tabular}} & \multicolumn{9}{c}{\bfseries Finetune Datasets} \\
\cmidrule(){2-10}
&\multicolumn{3}{c|}{ISLES15(DWI)}& \multicolumn{3}{c|}{ISLES15(FLAIR)}&\multicolumn{3}{c}{WMH(FLAIR)}\\
\midrule
Setting 1 &{\bfseries62.2} &60.0&59.1 & \multicolumn{3}{c|}{N/A}  &\multicolumn{3}{c}{N/A} \\
\midrule
Setting 2 &\multicolumn{3}{c|}{N/A}& {\bfseries58.2}& 55.1 & 53.4 & \bfseries{76.6}  &70.0& 69.6\\
\bottomrule
\end{tabular}
\end{table}

\section{Conclusion}
This study investigated the integration of an agnostic input channel alongside modality-specific input channels into a standard U-Net architecture. Thereby, providing a dedicated location to process an unseen modality at inference time in combination with modalities seen during training. We demonstrate that this channel can be trained in a modality-agnostic fashion by augmenting the inputs and employing a dedicated architecture centered on the agnostic channel, resulting in improved brain lesion segmentation across multiple settings consisting of various types of lesion and modality sets. 
This work not only demonstrates improved generalization but also provides evidence for domain adaptation to new domains with a combination of seen and previously unseen modalities. 

    

\begin{credits}
\subsubsection{\ackname} 
APA is supported by the Oxford-Bellhouse Graduate Scholarship, jointly funded by Magdalen College and the University of Oxford. FW is supported by the EPSRC Centre for Doctoral Training in Health Data Science (EP/S0242\- 8X/1), the Anglo-Austrian Society, and the Oxford-Reuben scholarship. NLV gratefully acknowledges support from the NIHR Oxford Health Biomedical Research Centre [NIHR203316]. The views expressed are those of the author(s) and not necessarily those of the NIHR or the Department of Health and Social Care. The Centre for Integrative Neuroimaging was supported by core funding from the Wellcome Trust (203139/Z/16/Z and 203139/A/16/Z).

\subsubsection{\discintname}
The authors have no competing interests.

\end{credits}

%
%
%

\bibliographystyle{splncs04}
\bibliography{references_cleaned}







\end{document}